\documentclass[10pt,twocolumn,letterpaper]{article}

\usepackage{iccv}
\usepackage{times}
\usepackage{epsfig}
\usepackage{graphicx}
\usepackage{amsmath}
\usepackage{amssymb}
\usepackage{tabularray}
\usepackage{rotating}
\usepackage{multirow}
\usepackage[accsupp]{axessibility}  % Improves PDF readability for those with disabilities.
\usepackage[pagebackref=true,breaklinks=true,letterpaper=true,colorlinks,bookmarks=false,linkcolor = blue,
            urlcolor  = blue,
            citecolor = blue,
            anchorcolor = blue]{hyperref}
\usepackage[breaklinks=true,bookmarks=false]{hyperref}

\iccvfinalcopy % *** Uncomment this line for the final submission

\def\iccvPaperID{****} % *** Enter the ICCV Paper ID here

\ificcvfinal\fi

\begin{document}

%%%%%%%%% TITLE
\title{NAPA-VQ: Neighborhood Aware Prototype Augmentation with Vector Quantization for Continual Learning}

\author{Tamasha Malepathirana \and Damith Senanayake \and Saman Halgamuge \\
Dept. of Mechanical Engineering\\
The University of Melbourne\\
{\tt\small \{tamasha.malepathirana, damith.senanayake, saman.halgamuge\}@unimelb.edu.au}
% \and
% Damith Senanayake\\
% Dept. of Mechanical Engineering\\
% The University of Melbourne\\
% % {\tt\small secondauthor@i2.org}
% \and
% Saman Halgamuge\\
% Dept. of Mechanical Engineering\\
% The University of Melbourne\\
}
\maketitle
\ificcvfinal\thispagestyle{empty}\fi

%%%%%%%%% ABSTRACT
\begin{abstract}
Catastrophic forgetting; the loss of old knowledge upon acquiring new knowledge, is a pitfall faced by deep neural networks in real-world applications. Many prevailing solutions to this problem rely on storing exemplars (previously encountered data), which may not be feasible in applications with memory limitations or privacy constraints. Therefore, the recent focus has been on Non-Exemplar based Class Incremental Learning (NECIL) where a model incrementally learns about new classes without using any past exemplars. However, due to the lack of old data, NECIL methods struggle to discriminate between old and new classes causing their feature representations to overlap. We propose NAPA-VQ: \textbf{N}eighborhood \textbf{A}ware \textbf{P}rototype \textbf{A}ugmentation with \textbf{V}ector \textbf{Q}uantization, a framework that reduces this class overlap in NECIL. We draw inspiration from Neural Gas to learn the topological relationships in the feature space, identifying the neighboring classes that are most likely to get confused with each other. This neighborhood information is utilized to enforce strong separation between the neighboring classes as well as to generate old class representative prototypes that can better aid in obtaining a  discriminative decision boundary between old and new classes. Our comprehensive experiments on CIFAR-100, TinyImageNet, and ImageNet-Subset demonstrate that NAPA-VQ outperforms the State-of-the-art NECIL methods by an average improvement of 5\%, 2\%, and 4\% in accuracy and 10\%, 3\%, and 9\% in forgetting respectively. Our code can be found in \url{https://github.com/TamashaM/NAPA-VQ.git}.
\end{abstract}

%%%%%%%%% BODY TEXT

\section{Introduction}

The achievements of deep neural networks over the years have grown significantly, and their efficiency and applicability have been demonstrated by numerous state-of-the-art works \cite{Krizhevsky2012ImageNetNetworks, He2016DeepRecognition, Long2015FullySegmentation, Girshick2015FastR-CNN}. However, a major requirement for the optimal operation of gradient-based optimization -- the ubiquitous learning paradigm -- is the data samples being independently and identically distributed (IID) \cite{Hadsell2020EmbracingNetworks}. This assumption may not hold in real data owing to various factors such as the addition of new classes of data, the removal of old data due to memory or availability constraints, and the changes in the data-generating phenomena (concept drift). As a result, neural networks may experience catastrophic forgetting where the network forgets the previously learned knowledge upon acquiring new knowledge \cite{Kirkpatrick2017OvercomingNetworks}. 

``Continual Learning'' is a field of research pursuing mechanisms to mitigate this forgetting \cite{DeLange2021ATasks}. In this manuscript, we focus on one paradigm of continual learning, named Class Incremental Learning (CIL) \cite{Lampert2017}. In CIL, a neural network is trained over a series of tasks and at each task, the network learns a new set of classes. At any given time, the network should classify between all learned classes thus far. %without having access to the task ID.
% CIL performs a task-agnostic evaluation \cite{Liu2020MoreLearning}, where the task identity of samples is not specified during evaluation. 
Among the techniques proposed for CIL, rehearsal-based methods have demonstrated promising results in mitigating forgetting by storing exemplars (old samples) and reusing them while learning new tasks \cite{Lampert2017, Castro2018End-to-endLearning,NIPS2017_f8752278}. However, such storage is not always possible due to memory limitations and privacy constraints \cite{Smith2022ALearning}. Therefore, we focus on Non-Exemplar based CIL (NECIL), a more pragmatic yet challenging scenario, which attempts to preserve the old knowledge without storing any exemplars \cite{Zhu2021PrototypeLearning, Zhu2022Self-SustainingLearning}. 

NECIL methods often struggle with overlapping old and new class representations due to the unavailability of exemplars, resulting in catastrophic forgetting. \cite{Zhu2021PrototypeLearning}.
While prototypes of old classes in the deep feature space are a viable alternative to reusing exemplars \cite{Zhu2021PrototypeLearning, Zhu2021Class-IncrementalAugmentation}, if not properly generated, the class boundaries refined using such prototypes tend to be muddled, causing confusion between the old and new classes, and in turn, leading to catastrophic forgetting. To overcome this limitation, we propose NAPA-VQ: Neighborhood-Aware Prototype Augmentation with Vector Quantization framework. NAPA-VQ not only proposes a novel way to create prototypes of old classes by considering class neighborhoods but also incorporates a novel quantization mechanism to create clearer class boundaries by enforcing a strong separation between the neighboring classes. This reduction in representation overlap effectively mitigates catastrophic forgetting. NAPA-VQ builds on the principles of unsupervised Neural Gas (NG) \cite{ThomasMartinetzandKlausSchulten1991ATopologies} and supervised Learning Vector Quantization (LVQ) \cite{Kohonen1990ImprovedQuantization} to facilitate this neighborhood awareness and enforce more discriminative boundaries. 

NAPA-VQ contains two  components: (I) a Neighborhood-aware Vector Quantizer (NA-VQ) and (II) a Neighborhood-aware Prototype Augmenter  (NA-PA).  NA-VQ learns the topology of the feature space manifold $\mathcal{Z}$ which is the output of a deep feature extractor (\eg ResNet), identifying the neighboring classes that share similar features and are hence prone to get confused with each other. This knowledge of the neighboring classes and their class distributions is utilized by NA-VQ to increase their separability and by NA-PA for old class prototype augmentation, i.e., to generate surrogate exemplars to facilitate the retention of old information when new classes are being learned.

To summarize,
\begin{itemize}
  \item We propose an improved supervised vector quantization method to discretize the latent space and improve class separation.
  \item We propose a prototype augmentation method that uses the topological information of classes in the latent space to avoid confusion between classes and catastrophic forgetting.
    \item We demonstrate the utility of the above two contributions combined in NECIL, obtaining superior performance compared to the existing NECIL methods on CIFAR-100, TinyImageNet, ImageNet-Subset, and ImageNet-1k datasets.
\end{itemize}

%-------------------------------------------------------------------------
\section{Related work}
\subsection{Incremental learning}

% Prior research in continual learning primarily centred around TIL, however, current attention has shifted towards the more challenging realm of CIL. 
% CIL presents a formidable challenge for the learner, as it must distinguish between classes from all tasks without having access to the task ID during inference.  
The techniques proposed to combat catastrophic forgetting can be broadly categorized into three \cite{DeLange2021ATasks}. 
% Although some of these works were originally proposed for Task Incremental Learning, they have also been adapted to CIL setting \cite{Masana2020}.
(1) The regularisation-based methods that add an extra regularisation loss term either to penalize changes to the network parameters that are important for previous tasks \cite{Kirkpatrick2017OvercomingNetworks, Liu2018RotateForgetting, Zenke2017ContinualIntelligence, Aljundi2018MemoryForget} or to distill knowledge from previous tasks to the current task \cite{Li2018LearningForgetting, Dhar2019LearningMemorizing,Zhang2020Class-incrementalConsolidation}. (2) The parameter isolation-based methods that assign each task with an isolated set of  parameters to prevent task interference either by dynamically increasing the network capacity \cite{Rusu2016ProgressiveNetworks,Yoon2017LifelongNetworks, Aljundi2017ExpertExperts} or by masking previous task parameters in a fixed size network \cite{Fernando2017Pathnet:Networks, Serra2018OvercomingTask,Mallya2018Packnet:Pruning}. Although parameter isolation methods are effective in overcoming catastrophic forgetting, they experience either a linear increase in network parameters or a decrease in capacity per task as the number of tasks grows \cite{Smith2021AlwaysLearning}. (3) The rehearsal-based methods that store a small subset of previous task data to either retrain \cite{Lampert2017,Castro2018End-to-endLearning,Chaudhry2019OnLearning} or constrain the optimisation \cite{NIPS2017_f8752278,Chaudhry2019EfficientA-gem,Aljundi2019GradientLearning} during the learning of new tasks in order to retain the discriminability between old and new classes. However, these methods also encounter pitfalls due to memory limitations, and other pragmatic concerns such as privacy or consent issues when storing samples.
An alternative to rehearsal-based methods is ``pseudo-rehearsal", which involves training a generative model to mimic past task distributions \cite{Shin2017,Seff2017ContinualNets}. Despite the encouraging results, generative models are computationally expensive to train \cite{deVen2018GenerativeLearning} and are also prone to catastrophic forgetting \cite{Thanh-Tung2020CatastrophicGans}. This motivated the development of NECIL strategies that neither depends on real nor fake past samples \cite{Zhu2021PrototypeLearning,Zhu2022Self-SustainingLearning,Liu2020MoreLearning,Yu2020}.

NECIL methods benefit from powerful feature extractors learning transferable features across tasks, as demonstrated by SDC \cite{Yu2020}, which showed that embedding networks suffer significantly less from catastrophic forgetting. PASS \cite{Zhu2021PrototypeLearning} also showed that self-supervised learning alleviates task-level overfitting. Furthermore, to maintain the decision boundaries of previously learned classes, PASS introduced a class-mean prototype augmentation technique based on Gaussian noise. While this technique aids in the retention of old information, it can be further improved by leveraging the knowledge of the distribution of classes in the feature space. Accordingly, IL2A \cite{Zhu2021Class-IncrementalAugmentation} proposed storing covariance matrices to retain class variations, but this approach can be memory intensive. SSRE \cite{Zhu2022Self-SustainingLearning} proposed a dynamic structure reorganization strategy to retain and transfer knowledge between tasks along with a prototype selection mechanism that utilizes an up-sampling technique of non-augmented class-mean prototypes. Similar to these approaches, we also store the mean prototype, while proposing a new method to augment them. To this end, we use the topological connections derived from an NG-like vector quantization to generate prototypes that lie within the shared feature regions of the confusing classes which aid in establishing better class discrimination.

\begin{figure*}[t]
\centering
\includegraphics[width=0.75\textwidth]{
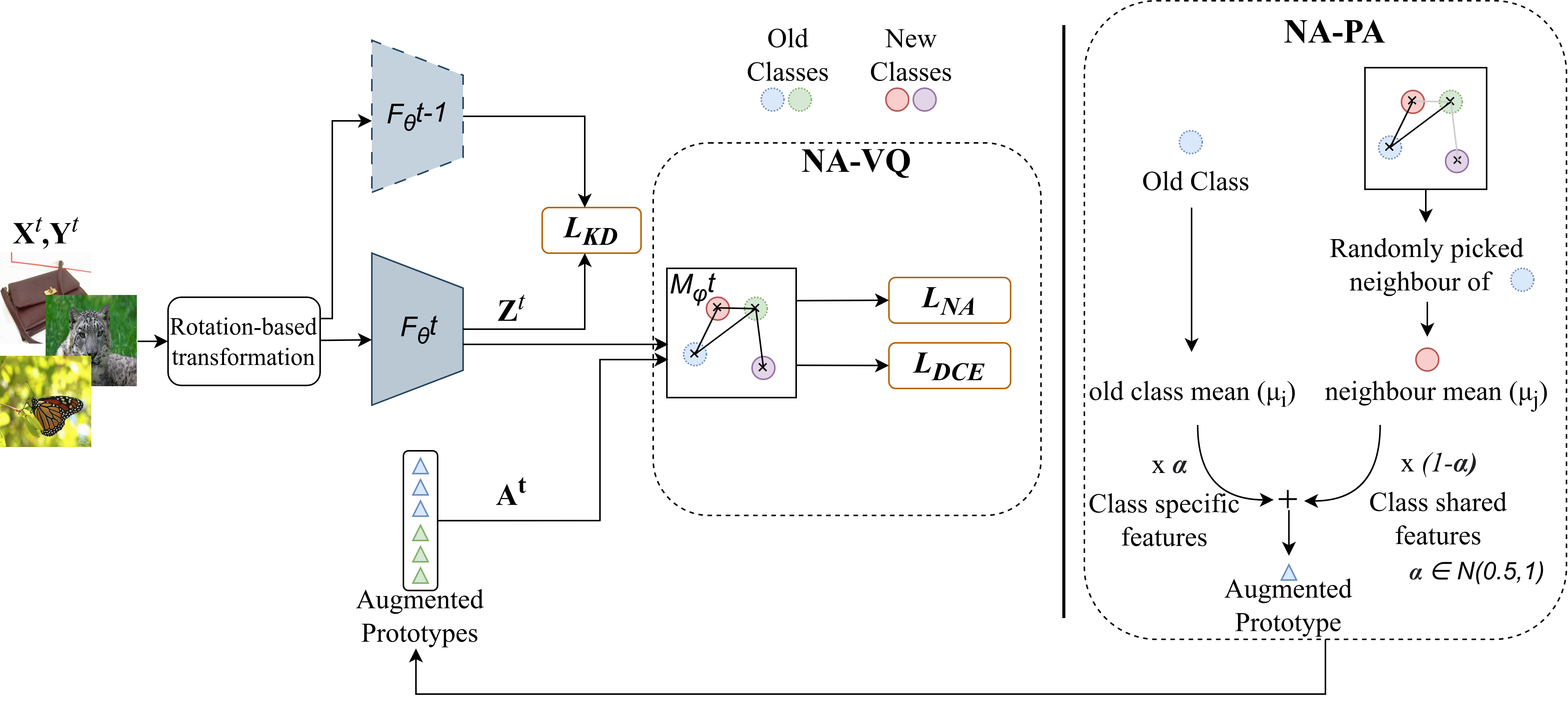}
\caption{Illustration of NAPA-VQ. Data from the current task $t$ are augmented using a rotation-based technique \cite{Zhu2021PrototypeLearning} and are fed to the feature extractor.
The obtained feature representations ($\mathbf{Z}^t$) and the NA-PA generated old class representative prototypes ($\mathbf{A}^t$) are sent to the vector quantizer (NA-VQ) to identify and repel confusing classes, establishing better discrimination in the feature space. 
% NA-VQ uses the generated old class representative prototypes ($A^t$) alongside the obtained feature representations ($Z^t$) to reducing the class overlap.
Knowledge Distillation ($L_{KD}$) is used to minimize the feature drift across tasks.}\label{fig:framework}
\end{figure*} 

\subsection{Vector quantization}

%Competitive Learning Networks are a category of neural network architectures, where neurons compete with each other for activation \cite{Kohonen1990TheMap}.
Vector Quantization (VQ), a technique used to discretize a continuous data space into a finite set of ``coding vectors'' (CVs) was popularised with the advents of Self-organizing Maps (SOMs) \cite{Kohonen1990TheMap}. In addition to quantizing the data manifold, a SOM captures a topological mapping from data to the CVs. Neural Gas (NG) networks \cite{ThomasMartinetzandKlausSchulten1991ATopologies,Fritzke1994ATopologies}, on the other hand, were introduced to address a shortcoming of the original SOM by allowing a generic graph structure rather than a fixed lattice structure. In NG, the CVs are adjusted to capture the data-dense regions, and the edges between these CVs are formed based on their proximity. These edges and the CVs form a graph that approximates the topology of the data manifold. % CHL develops topological connections between CVs that closely resemble the topology of the data manifold and NG-quantisation adjusts the CVs to ensure they reside within the data manifold.

% Concretely, CHL learns a topological structure that closely resembles the topology of the data by developing edges between the CVs that reside within the data manifold or its vicinity. Moreover, 

Coding vector-based learning can be traced back to the K-nearest neighbor (K-NN) algorithm \cite{Fix1989DiscriminatoryProperties}. 
%The nearest-neighbour (1-NN) rule, a special case of the K-NN rule, classifies the unknown pattern to the class of the nearest CV \cite{Cover1967NearestClassification}.
For instance, Learning Vector Quantization (LVQ) was proposed to derive the CVs used in a 1-NN classifier \cite{Kohonen1990ImprovedQuantization, Sato1996GeneralizedQuantization}.
Despite common roots, LVQ algorithms and unsupervised VQ algorithms such as SOM differ in their primary usages of CVs; the unsupervised algorithms attempt to obtain a set of CVs to best represent the data while LVQ algorithms attempt to reduce the misclassification rate by focusing on the decision boundaries between classes. These complementary properties allow us to combine unsupervised and supervised VQ methods \cite{Hammer2005SupervisedMeasure,DeVries2016DeepQuantization} to obtain CVs to both reduce the misclassification rate and represent the data distribution \cite{Kohonen1990TheMap}.
%Secondly, LVQ algorithms do not consider the relative positioning of the CVs whereas algorithms such as SOM use a neighbourhood function to adjust the CVs that are close to the winner CV ensuring that the CVs are well-spread out in the feature space forming a topological map of the input data. Therefore, by combining LVQ-based methods with unsupervised competitive learning methods, it is possible to obtain CVs that not only reduce the misclassification rate but also represent the data distribution well \cite{Kohonen1990TheMap}. This allows for a more accurate classification of new data points. As a result, the combination of LVQ with NG termed Supervised Neural Gas was investigated \cite{Hammer2005SupervisedMeasure,DeVries2016DeepQuantization} which demonstrated stable and robust behaviour. However, in the above-supervised algorithms, the vector quantisation process did not depend on the topology generated by CHL thus CHL was optional. However, CHL is useful in direct adaptations in the local neighbourhood as was shown in the NG extensions \cite{Fritzke1995ATopologies}. Therefore in our Supervised Neural Gas version, we utilise the topology formed in CHL in the vector quantisation step. 

% [cite]. It is worth exploring the advantages of this approach for continual learning.

Multiple studies explored the integration of the hierarchical feature-extracting capability of deep feature extractors with VQ \cite{DeVries2016DeepQuantization, Villmann2017FusionLearning,Blaes2017Few-shotPrototyping} which were also later adapted to Continual Learning. TPCIL \cite{Tao2020Topology-PreservingClassificatio} proposed to retain the topology of the feature space to preserve old knowledge over the increments. IDLVQ \cite{Chen2021IncrementalSpace} proposed to adapt a margin-based loss for the task of few-shot class incremental learning (FSCIL) -- a special case of CIL, therefore not directly transferrable to CIL/NECIL -- to create a large margin between classes to mitigate overlap. TOPIC \cite{Tao2020Few-ShotLearning} was also proposed for the FSCIL setting with the aim of preserving old knowledge by stabilizing a NG network. 
% Further, given an input, TOPIC updates all the CVs similar to \cite{10Topologies}. %However, we perform a localized update where only the direct neighbours of the correct CV (as determined by CHL) get updated avoiding unnecessary adjustments to CVs. 
We highlight that changes to the topology are possible due to the inevitable feature drift occurring over incremental steps thus a method that uses both augmented prototypes and new data to update the topological graph between CVs is warrented.% propose to update the topological graph between CVs using both the augmented prototypes and the new data.

\section{Methodology}

\subsection{Preliminaries and notations}
In CIL, a model is continually trained over a series of tasks, where at each task the model learns a set of new classes that are distinct from the previously learned classes. At any given time, the model should classify between samples from all classes seen thus far. The training data at task $t$ is denoted as $\mathbf{D}^{t} = \{\mathbf{X}^{t}, \mathbf{Y}^{t}\}$ 
% consistent with previous works\cite{Liu2020MoreLearning} 
where $\mathbf{X}^{t} = \{x^t_i\}_{i=1}^{N^t}$ is the set of input images and $\mathbf{Y}^{t} = \{y^t_i| y^t_i \in C^t\}_{i=1}^{N^t}$  are their target labels. ${N^t}$ and $C^t$ correspond to the number of samples and the set of classes at task $t$. We further define $ P^t = \sum_{j=1}^{t}|C^j|$ as the total number of classes seen at the end of task $t$.

\subsection{Overview of the framework}

An illustration of our framework is shown in Fig. \ref{fig:framework} which consists of a feature extractor ($F_{\theta}$), a Vector Quantizer (NA-VQ), and a prototype augmenter (NA-PA). $F_{\theta}$ is used to obtain the feature space of the input data. NA-VQ quantizes this feature space by learning a set of CVs named $M_{\phi}$ such that $m_i$ is associated with class $i$. These CVs are trained to effectively improve the inter-class variance and reduce intra-class variance in the feature space reducing the representational overlap in classes.
% = \{m_i \in R^n| i = 1,..., P^t\}$ such that $m_i$ encodes the feature representations of class $i$.
% We learn one CV per class thus 
% The classification is conducted using the nearest class CV rule in the feature space using Euclidean distance ($d$).
The module parameters $\theta$ and $\phi$ are shared across all tasks but are updated continually with the data at the current task, thus $t$ denotes the states of the parameters at each task.
Consequently, $F_{\theta^t}$ and $M_{\phi^t}$ refer to the states of the feature extractor and the set of CVs at task $t$.
% thus superscript $t,i$ is used to denote the parameter state. For example, $F_{\theta^{t,i}}$ and $M_{\phi^{t,i}}$ refer to the module state after training with the $i^{th}$ sample of task $t$. However, we drop the second index $i$ and use $F_{\theta^t}$ and $M_{\phi^t}$ to refer to the module state at any point in task $t$.  
 The goal of task $t$ is to jointly update $F_{\theta^{t-1}}$ and $M_{\phi^{t-1}}$ using $\mathbf{D}^{t}$ to obtain $F_{\theta^t}$ and $M_{\phi^t}$. At incremental tasks (when $t>0$), a set of old class prototypes ($\mathbf{A}^t$) generated using our NA-PA technique is used alongside $\mathbf{D}^t$ to retain the discrimination between old and new classes.

\begin{figure}[t]
\centering
\includegraphics[width=0.4\textwidth]{
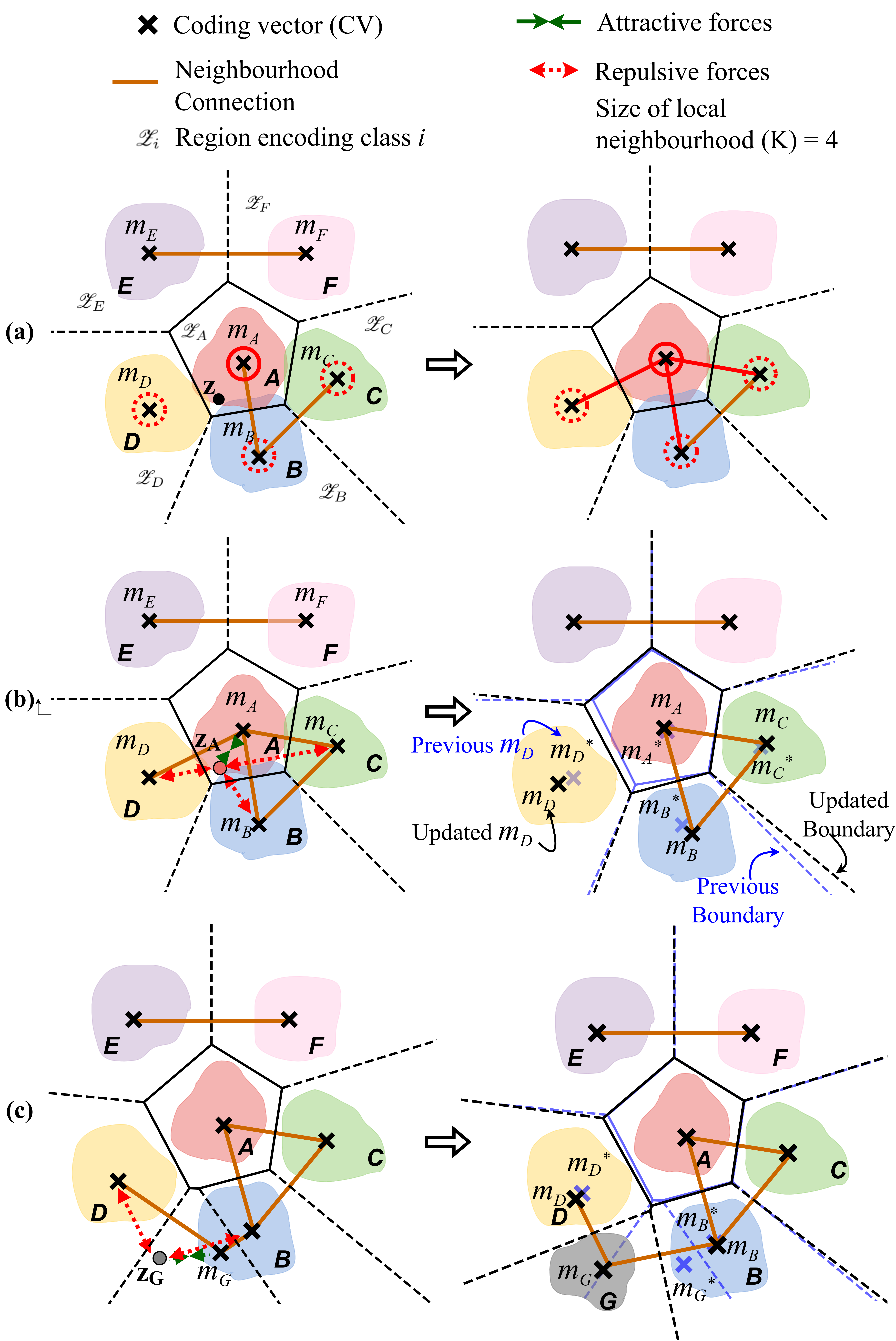}
\caption{Conceptual illustration of NA-VQ. 
% Each subregion $\mathscr{Z}_i$ represented by $m_i$ encodes the feature representations of class $i$. 
Each color represents a single class. 
% Neighbourhood connections between the CVs represent the topological relationships between the classes. 
(a) Given $z$, edges are created between the closest CV (solid circle) and the next $K-1$ closest CVs (dotted circles) to approximate the local topology around $z$. (b) Class A is overlapped with classes B, C, and D thus a sample from class A ($z_A$) could be misclassified into B, C, or D. NA-VQ pulls $z_A$ and $m_A$ together and pushes $z_A$ and direct neighbors of $m_A$ ($m_B,m_C$, and $m_D$) away from each other, reducing the class overlap. Closer the CV to $z_A$, the higher these forces. 
% $m_i^*$ corresponds to the previous position on $m_i$. 
The edge between $m_A$ and $m_D$ is pruned due to edge weakening over time. (c) When a new class G is introduced, $m_G$ is inserted randomly and refined through attractive and repulsive forces over the iterations, identifying a less-overlapping space in the feature space.}
 \label{fig:concept}
\end{figure}
% $\theta$ and $\phi$ are trained jointly in an end-to-end manner.
% \footnote{Hereafter, we remove the superscript $t$ and subscript $i$ for easy demonstration.}.
% to make them cooperate better with each other, which is beneficial for the performance of classification.

\subsection{Neighborhood-Aware Vector Quantizer (NA-VQ)}\label{sec:navq}

% At the beginning of task $t$, one CV per new class ($|C^t|$ CVs in total) is randomly initialized and added into $M_{\phi^{t-1}}$. 

At task $t$, we extract the features of $\mathbf{X}^t$ by computing $\mathbf{Z}^t \subset \mathbb{R}^n $ using $F_{\theta^t}$, where $n$ is the dimensionality of the feature space (Eq. \ref{eq:ft}). 
\begin{align}
\mathbf{Z}^t = F_{\theta^t}(\mathbf{X}^t)
\label{eq:ft}
\end{align}

We assume $\mathbf{Z}^t$ lies in a feature space manifold $\mathcal{Z}$ which captures the topological properties between the classes. NA-VQ partitions $\mathcal{Z}$ into disjoint regions such that $\mathcal{Z}_i$ contains the feature distribution of class $i$. We encode $\mathcal{Z}_i$ using $m_i$, i.e., $M_{\phi^t} = \{m_i \in \mathbb{R}^n| i = 1,..., P^t\}$ contains one CV per class.
 The algorithm can be summarised into two iterative steps (1) the topology approximation and (2) the CV adaptation. 
% In contrast to the previous versions supervised-NG \cite{ Hammer2005SupervisedMeasure, DeVries2016DeepQuantization},
These steps are performed concurrently, i.e., the topology learned depends on the CV adaptation and vice versa. A conceptual illustration of NA-VQ is shown in Fig. \ref{fig:concept}.

\textbf{(1) Topology approximation.}
Inspired by NG \cite{ThomasMartinetzandKlausSchulten1991ATopologies}, the topology of $\mathcal{Z}$ is approximated by learning an undirected graph $G^t = <M_{\phi^t}, E^t>$, where CVs are the nodes and the topological connections between the CVs are the edges represented by the adjacency matrix $E$. 
At the beginning of task $t$, $M_{\phi^{t-1}}$ and $E^{t-1}$ are extended to $M_{\phi^{t}}$ and $E^{t}$ to accomodate task $t$. Specifically, $C^t$ new randomly initialized CVs are inserted without any edges that link to, from, or between them.
 During the learning of task $t$, $E^{t}$ is modified by adding, decaying, and removing edges between the CVs. 

% Only the CVs that are located in the vicinity of $z \in Z^t$ establish any edges, hence the term ``local density information". 

Concretely, given $(x,y) \in (\mathbf{X}^t, \mathbf{Y}^t)$ and its feature representation $z \in \mathbf{Z}^t$, we calculate the distance from $z$ to each CV in $M_{\phi^{t}}$:
$D = \{ d(z, m_i) | i = 1,...,P^t\} $ where $d$ is the Euclidean distance. 
$D$ is then sorted to assign a rank to each CV:
$R = \{ r_i | i= 1,...,P^t\}$ such that $\forall i, d(z, m_{r_i})<d(z, m_{r_{i+1}})$. Next, the edges between the CVs are formed based on a connectivity factor denoted as $K$. A higher value of $K$ leads to a denser graph, whereas a lower value of $K$ results in a sparser graph. Specifically, edges are created between the closest CV and the next $K-1$ closest CVs, i.e., $e_{{r_1},j}\gets1$ for $j = r_{2},...,r{_{K}}$. 
% We note that the closest CV identified here may or may not be the correct CV of $x$, i.e., $m_y$.
Consequently, CVs that lie within the high-density regions of $\mathcal{Z}$  develop edges, allowing to identify classes that share similar features in $\mathcal{Z}$. Since the CVs get updated over time, the edges created in a previous iteration may become obsolete as their endpoints may have moved. To remove such edges we employ an edge decaying mechanism. All edges from the closest CV are decayed by a constant multiplier $\epsilon \in (0,1)$ so that the edges created at a previous iteration that no longer fall within a high-density region are weakened, i.e., $e_{r_{1},j}\gets e_{r_{1},j} * \epsilon$ for $ j = 1,.., P^t $.
% \begin{equation}
% e_{r_{1},j}= e_{r_{1},j} * \epsilon, j = 1,.., P^t
% \end{equation}
If the edge strength goes below a predefined $e_{\mathrm{min}}$, such edges would be pruned, i.e., if $e_{i,j} < e_{\mathrm{min}}$, $e_{i,j} \gets 0$.
% $e_{i,j}= 0 \text{ for } e_{i,j} < e_{min}, i = 1,..., P^t \text{ and } j = 1,..., P^t$
%, thus this step follows an unsupervised topology approximation. 

% At task $t$, given an input sample $<x,y> \in D^t$, we obtain its feature representation $z$ simply by forwarding it through $F_{\theta^t}$;
% \begin{equation}
% z = F_{\theta^t}(x)
% \end{equation}
% Then the probability of $z$ belonging to the feature region represented by $m_i$ can be measured by the Euclidean distance ($d$) between $z$ and $m_i$ \cite{Yang2018RobustLearning}.
% \begin{equation}
% p(z \in m_i |x) \propto -d(z,m_i)
% \end{equation}
% To satisfy the non-negative and sum-to-one properties of the probability, we further define $p(z \in m_i |x)$ as the equation below where $\tau$ is a temperature parameter.
% \begin{equation}
% p(z \in m_i |x)=\frac{e^{-\frac{||z-m_i||}{\tau} }}{\sum_{k=1}^{P^t}e^{-\frac{||z-m_k||}{\tau}}}
% \end{equation}
% Consequently, the probability of sample $x$ belonging to class $y$, $p(y|x)$ can be defined by the probability of $z$ corresponding to $m_y$.
% \begin{equation}
% p(y|x) =p(z \in m_y |x) 
% \end{equation}
% The distance-based cross-entropy loss  ($L_{DCE}$) for $X^t$ can then be defined as below \cite{Yang2018RobustLearning} where $Z^t = F_{\theta^t}(X^t)$.
% \begin{equation}
% L_{DCE}^t(<Z^t,Y^t>; M_{\phi^t}) = \sum_{<z,y> \in <Z^t,Y^t>}-\log p(z \in m_y |x) 
% \label{eq:dce}
% \end{equation}

%CHL should I add?
\textbf{ (2) CV adaptation.} During the learning of task $t$, the class label $y$ in $(x,y) \in (\mathbf{X}^t, \mathbf{Y}^t)$ is employed to adapt CVs in  {$M_{\phi^{t}}$} to improve the discriminability in the feature space. Concretely, we create attractive forces between $z$ and $m_y$ and repulsive forces between $z$ and any other ``confusing CVs" of $m_y$. The confusing CVs of $m_y$ (referred to as $N_{y}^-$) are the direct neighbors of $m_y$ as determined by graph $G^t$ in Step (1). 
\begin{align}
N_{y}^-= \{ m_i |e_{y,i}>0 \text{ and } y \neq i \text{ and } i= 1,...,P^t \} 
\end{align}
Any $m_i \in N_{y}^-$ is considered to be close to $m_y$, thus, there is a high likelihood that $m_i$ will be mistaken for $m_y$ as the winner CV when presented with sample $x$. To counteract this, we propose a Neighborhood-Adaptation loss ($L_{NA}$) (Eq. \ref{eq:sung}) to bring $z$ and $m_y$ closer together while pushing $z$ and $N_{y}^-$ farther apart. 
\begin{align}
L_{NA}^t(\mathbf{X}^t,\mathbf{Y}^t; M_{\phi^t},F_{\theta^t}))=\sum_{\substack{{x,y} \in \\ \mathbf{X}^t,\mathbf{Y}^t}} ReLU(d(z, m_y)-d_{\mathrm{neigh}-})  \nonumber \\ 
\text{ where } d_{\mathrm{neigh}-} = \sum_{m_i \in N_{y}^- }^{}W_{m_i} \times d(z,m_i) \text{ and } z=F_{\theta^t}(x)
\label{eq:sung}
\end{align}
$L_{NA}$ computes the difference between $d(z,m_y)$ and $d_{\mathrm{neigh}-}$ when $d(z,m_y) > d_{\mathrm{neigh}-}$, the scenario where $x$ is most likely to be misclassified.
$d(z,m_y)$ is the distance from $z$ to its correct CV $m_y$ and $d_{\mathrm{neigh}-}$ is a linear combination of the distances from $z$ to $N_{y}^-$. The weight given to $m_i \in N_{y}^-$ ($W_{m_i}$) is calculated using a monotonic decaying function (Eq. \ref{eq: wm}) to reduce the impact of $m_i$ on $L_{NA}$ as the distance between $m_i$ and $z$ increases. 

 \begin{equation}
W_{m_i} = \frac{e^{-\beta \times d(z,m_i)}}{\sum_{m_j \in N_{y}^-}^{}e^{-\beta \times d(z,m_j)}} \text{ where } \beta >0 
\label{eq: wm}
\end{equation}

To avoid the repelling forces in $L_{NA}$ from diverging any $m_i \in N_{y}^-$ from their respective class distributions, we take three steps. First, the repelling forces are distributed over multiple confusing CVs. Second, $ReLU$ prevents any adjustments to the CVs when $d(z,m_y) < d_{\mathrm{neigh}-}$, since it is unlikely that this scenario would lead to misclassification. Finally, Distance-based cross-entropy loss ($L_{DCE}$) \cite{Yang2018RobustLearning}, as described next, is used to encourage $m_i$ to more accurately represent the distribution of class $i$ which aids in minimizing the distance between the feature representations of class $i$ and $m_i$. 

The probability of sample $x$, belonging to class $y$ can be measured by the distance between $z = F_{\theta^t}(x)$ and $m_y$ \cite{Yang2018RobustLearning}, i.e., $p(y|x) \propto -d(z,m_y)$. 
% \begin{equation}
% p(y|x) \propto -d(z,m_y)
% \end{equation}
Considering the non-negative and sum-to-one properties of the probability, we can define $p(y|x)$ using a softmax function as shown in Eq. \ref{eq:dce_sub} where $\tau$ is a temperature parameter.
\begin{align}
p(y|x)=\frac{e^{-\frac{d(z,m_y)}{\tau} }}{\sum_{j=1}^{P^t}e^{-\frac{d(z,m_j)}{\tau}}}
\label{eq:dce_sub}
\end{align}
% Consequently, the probability of sample $x$ belonging to class $y$, $p(y|x)$ can be defined by the probability of $z$ corresponding to $m_y$.
% \begin{equation}
% p(y|x) =p(z \in m_y |x) 
% \end{equation}
Since the true probability distribution is a one hot encoded vector,  we can define $L_{DCE}$ for $\mathbf{X}^t$ as Eq. \ref{eq:dce}.
\begin{equation}
L_{DCE}^t(\mathbf{X}^t,\mathbf{Y}^t; M_{\phi^t}, F_{\theta^t}) = \sum_{x,y \in \mathbf{X}^t,\mathbf{Y}^t}-\log p(y|x) 
\label{eq:dce}
\end{equation}

These two steps are conducted concurrently at each mini-batch gradient update during the optimization of $\theta$ and $\phi$. By backpropagating the gradients calculated for $L_{NA}$ and $L_{DCE}$ through $F_\theta$ and $M_\phi$, we effectively reduce the class overlapping in the feature space and establish a more discriminative decision boundary between classes. During the incremental steps $(t>0)$, we freeze the old class representative CVs but update their topological connections using both the augmented prototypes (Sec. \ref{sec:pa}) and $\mathbf{D}^t$.

\subsection{Neighborhood-Aware Prototype Augmenter (NA-PA)}\label{sec:pa}

In NECIL, we cannot directly 
% optimize the network parameters to maintain the differentiation between the old classes or establish a differentiation between the old and new classes since we cannot 
compute $L_{NA}$ or $L_{DCE}$ for the previous task samples. 
% Instead, we store class representative mean prototypes in the feature space. 
While the class mean in the feature space serves as a central representative point for each old class, it does not capture the class variance. Thus, if the model relies solely on class-means without any augmentation, it may overfit to the class mean and forget to distinguish samples of the same class that are close to but not exactly equal to the class mean. Augmentation mitigates this issue by generating multiple representative points for each old class, considering the underlying class variance. Therefore, we store representative mean prototypes
and augment them on the fly with NA-PA during the learning of each new task. Specifically, we generate prototypes in the regions where the model is uncertain about the class labels. We consider pairs of classes that the model may confuse between using the topological graph $G^t$ established in Sec. \ref{sec:navq}, and generate prototypes to distinguish between these pairs of classes, increasing the quality of the prototypes used for retaining old knowledge.

Inspired by the work \cite{Chu2020FeatureData}, we identify that the features of each class can be decomposed into a class-specific component (features placed closer to the class mean) and a class-shared component (features that lie between itself and another class). For a given old class, the samples near the boundaries shared with its confusing classes have the biggest impact on recovering good decision boundaries because they are closer to the regions where the model is uncertain about the class labels. Thus, NA-PA generates augmented representations of the old classes by fusing the class-specific features from the old classes with the class-generic/shared features from their confusing classes to create such high-impact prototypes (Fig. \ref{fig:framework}). 
% Fortunately, we can identify the confusing classes by the direct topological neighbours of a particular class from $G^t$ without performing additional computations. 
At task $t(>0)$, an augmented prototype $a_i$ of old class $i$ is generated as shown in Eq. \ref{eq:ai} by fusing the mean prototype of class $i$ ($\mu_i$) with the mean prototype of class $j$  ($\mu_j$), a randomly picked neighbor of class $i$. By varying $\alpha$, we ensure the augmented prototypes are composed with varying degrees of uncertainty, i.e., the lower the $\alpha$, the higher the uncertainty. Here, we are using the mean prototypes instead of CVs as these CVs are adjusted continually to reduce misclassification and may potentially be positioned near class boundaries (away from class means) for improved discrimination. We determined that class-means serve as better representatives of the class-specific features than the CVs. 

\begin{align}
    a_i = \alpha \mu_i + (1-\alpha)\mu_j \text{ where } m_j\in N_{i}^- \text{ \& } \alpha \sim \mathcal{N}(0.5,1)
\label{eq:ai}
\end{align}
$\mathbf{A}^t$ is the collection of augmented prototypes and their class labels representing all old classes up to task $t$. 
% where $L_i$ is the number of prototypes from class $i$;
% \begin{equation}
%     \mathbf{A}^t = \bigcup_{i=1}^{P^{t-1}}\{a_{k,i},i\}_{k=1}^{k=L_i}
% \end{equation}
$\mathbf{A}^t$ is used alongside $\mathbf{D}^t$ in NA-VQ for topology approximation as well CV adaptation. Specifically, we calculate $\Hat{L}_{NA}$ and $\Hat{L}_{DCE}$ using $\mathbf{A}^t$. 
\begin{eqnarray}
    \Hat{L}_{DCE}^t(\mathbf{A}^t; M_{\phi^t}) = \sum_{a_i \in \mathbf{A}^t}-\log p(i|a_i) 
\\
\Hat{L}_{NA}^t(\mathbf{A}^t; M_{\phi^t})) =  \sum_{a_i \in \mathbf{A}^t} ReLU(d(a_i,m_i)-d_{\mathrm{neigh}-})
\end{eqnarray}

\subsection{Knowledge distillation}
% When training a model continually, Knowledge Distillation \cite{Li2018LearningForgetting, RebuffiICaRL:Learning,Zhu2021PrototypeLearning, Zhu2021Class-IncrementalAugmentation}
% the key to mitigating forgetting when learning a new task is to transfer knowledge from the most recent checkpoint into the current model.
% two strategies, namely prediction distillation  and feature distillation \cite{}. 
As $F_\theta^{t}$ gets updated continually, the actual feature distributions of old classes drift away from their original distributions. To mitigate this drift we incorporate a feature-level knowledge distillation ($L_{KD}$) \cite{Zhu2021PrototypeLearning, Zhu2021Class-IncrementalAugmentation} that attempts to align the feature spaces of the current and  the previous models.
 \begin{equation}
L_{KD}^t(\mathbf{X}^t|F_\theta^{t-1},F_\theta^t) = d(F_\theta^{t-1}(\mathbf{X}^t),F_\theta^t(\mathbf{X}^t))
\label{eq:kd}
 \end{equation}
The total loss used in our framework is shown below and $\lambda_1$ and $\lambda_2$ are loss weights (See Supp. Materials for explanations on loss weights)
\begin{equation}
L_{total}^t = L_{DCE}^t + L_{NA}^t \text{ when $t=0$ }
\end{equation}
\begin{equation}
L_{total}^t = L_{DCE}^t + \lambda_1 \Hat{L}_{DCE}^t + 
L_{NA}^t+\Hat{L}_{NA}^t+ \lambda_2 L_{KD}^t 
\text{ when $t>0$ } \nonumber
\end{equation}

\subsection{Rotation-based data augmentation}
% SSL has previously shown remarkable performance in continual learning \cite{Zhang2020Self-SupervisedLearning, Gallardo2021Self-SupervisedLearning, Zhu2021PrototypeLearning}. 
In order to learn richer features, we transform the training data using the same rotation-based approach used in \cite{Zhu2021PrototypeLearning}. Concretely, the training samples are rotated by 90, 180 and 270 degrees to generate 3 new pseudo-classes, learning $4P^t$ classes instead of $P^t$ classes at the training stage. However, the classification occurs only between the original classes during the evaluation stage.

\subsection{Classification}
We perform the nearest CV-based classification. At the end of task $t$, given a test sample $x$, we obtain $z= F_{\theta^t}(x)$, calculate the distance from normalized $z$ to each normalized CV and assign the class label of the closest CV to $x$. 
\begin{equation}
    y_{pred} =  \underset{i=1,...,P^t }{\arg\min}d(\frac{z}{||z||},\frac{m_i}{||m_i||})
\end{equation}

\section{Experiments}

\subsection{Datasets}
We perform comprehensive experiments using four datasets; CIFAR-100 \cite{Krizhevsky2009LearningImages}, TinyImageNet \cite{Le2015TinyChallenge}, ImageNet-Subset \cite{Deng2009Imagenet:Database}, ImageNet-1K \cite{Deng2009Imagenet:Database} in three incremental scenarios; $T=5, 10 \text{ and } 20$ where $T$ is the number incremental tasks. For comparability, the classes are arranged into tasks using the same fixed random order and division settings as \cite{Zhu2021PrototypeLearning, Zhu2022Self-SustainingLearning} for the first three datasets. For the ImageNet-1K dataset, we train the model on 400 classes for the first task, and equal classes in the rest of the tasks. 
% After each task, the model is evaluated on all the learned classes so far. 
% For a fair comparison, the classes are arranged into tasks using the same fixed random order as \cite{Zhu2021PrototypeLearning, Zhu2022Self-SustainingLearning}.  

\subsection{Implementation details} \label{sec:impl}
For a fair comparison, we adapted the same backbone architecture, ResNet-18 \cite{He2016DeepRecognition} from \cite{Zhu2021PrototypeLearning}. The concrete details of our implementation can be found in the Supp. materials and the \href{https://github.com/TamashaM/NAPA-VQ.git}{publicly available codebase}.
All the experiments were conducted on the University of Melbourne’s high-performance computing system, Spartan \cite{Lafayette2016SpartanChimera}.
% We use two optimizers, $Optim_{\theta}$ and $Optim_{\phi}$ to jointly train the $F_\theta$ and $M_\phi$. 
% For CIFAR-100 and TinyImageNet, an Adam optimizer was used as the $Optim_{\theta}$ and an SGD optimizer was used as the $Optim_{\phi}$ whereas for ImageNet-Subset, SGD optimizers were used for both $Optim_{\theta}$ and $Optim_{\phi}$. 
% More details related to training can be found in supplementary materials.

% An Adam optimizer was used as the $Optim_{\theta}$ and an SGD optimizer was used as the $Optim_{\phi}$. The base task was trained with an initial learning rate (lr) of 0.001 for $Optim_{\theta}$ and 5 for $Optim_{\phi}$. The lr decayed by 0.1 every 45 epochs for  $Optim_{\theta}$ and every 20 epochs for $Optim_{\phi}$.  For ImageNet-Subset, the base task was trained for 160 epochs whereas the incremental tasks were trained for 100 epochs. SGD optimizers were used for both $Optim_{\theta}$ and $Optim_{\phi}$. The base task was trained with an initial lr of 0.1 for $Optim_{\theta}$ and 5 for $Optim_{\phi}$. The lr decayed by 0.1 at steps 80, 120, and 150 for $Optim_{\theta}$ and by 0.1 every 20 epochs for $Optim_{\phi}$. At incremental steps in all datasets, we initiate with a lower lr for both optimisers (0.0001 for $Optim_{\theta}$ and 0.5 for $Optim_{\phi}$), however, the lr decayed by 0.1 every 45 epochs for  $Optim_{\theta}$ and every 20 epochs for $Optim_{\phi}$. Parameter values of $\tau =10,K=15,\epsilon=0.9, e_{min} =0.9^{10},\alpha=0.001 $ and loss weights of $ \lambda_1=10$ and $\lambda_2=10$ were used for all three datasets.
\begin{table*}
\centering
\caption{Average Accuracy (\%) of NAPA-VQ compared to the top three SOTA using four datasets. The higher the values, the better. $T$ is the number of incremental tasks. Values for the methods with * were extracted from \cite{Zhu2022Self-SustainingLearning}. Our improvement is shown in red.}.%except for the $T=5 \& 20$ which were reproduced from their original code repository.Our improvement is shown in red.}
\label{Tab:accuracy}
\begin{tblr}{
  width = \linewidth,
  colspec = {Q[98]Q[70]Q[70]Q[70]Q[70]Q[70]Q[70]Q[70]Q[70]Q[70]Q[70]Q[70]Q[70]},
  cell{1}{1} = {r=2}{},
  cell{1}{2} = {c=3}{0.200\linewidth},
  cell{1}{5} = {c=3}{0.200\linewidth},
  cell{1}{8} = {c=3}{0.200\linewidth},
  cell{1}{11} = {c=3}{0.200\linewidth},
  vlines,
  hline{1,3-7} = {-}{},
  hline{2} = {2-13}{},
}
Method           & CIFAR-100                                           &                                                  &                                                     & TinyImageNet                                        &                                                     &                                                     & ImageNet-Subset                                     &                                                     &                                                     & ImageNet-1K &      &      \\
                 & T=5                                                 & T=10                                             & T=20                                                & T=5                                                 & T=10                                                & T=20                                                & T=5                                                 & T=10                                                & T=20                                                & T=5         & T=10 & T=20 \\
PASS *           & 63.47                                               & 61.84                                            & 58.09                                               & 49.55                                               & 47.29                                               & 42.07                                               & 66.84                                               & 61.80                                               & 54.46                                               & -           & -    & -    \\
IL2A             & 65.61                                               & 59.09                                            & 58.82                                               & 47.02                                               & 44.48                                               & 39.68                                               & -                                                   & -                                                   & -                                                   & -           & -    & -    \\
SSRE *           & 65.88                                               & 65.04                                            & 61.70                                               & 50.39                                               & 48.93                                               & 48.17                                               & -                                                   & 67.69                                               & -                                                   & -           & -    & -    \\
\textbf{NAPA-VQ} & {\textbf{70.44}\\\textbf{\textcolor{red}{\small{(+4.56)}}}} & {\textbf{69.04}\\\textbf{\textcolor{red}{\small{(+4)}}}} & {\textbf{67.42}\\\textbf{\textcolor{red}{\small{(+5.72)}}}} & {\textbf{52.77}\\\textbf{\textcolor{red}{\small{(+2.38)}}}} & {\textbf{51.78}\\\textbf{\textcolor{red}{\small{(+2.85)}}}} & {\textbf{49.51}\\\textbf{\textcolor{red}{\small{(+1.34)}}}} & {\textbf{69.15}\\\textbf{\textcolor{red}{\small{(+2.31)}}}} & {\textbf{68.83}\\\textbf{\textcolor{red}{\small{(+1.14)}}}} & {\textbf{63.09}\\\textbf{\textcolor{red}{\small{(+8.63)}}}} & \textbf{55.11}        & \textbf{53.04} & \textbf{45.46} 
\end{tblr}
\end{table*}
\begin{table*}
\centering
\caption{Average Forgetting (\%) of NAPA-VQ compared to the top three SOTA using four datasets. The lower the values, the better. $T$ is the number of incremental tasks. Values for the methods with * were extracted from \cite{Zhu2022Self-SustainingLearning}. Our improvement is shown in red.}
\label{Tab:forgetting}
\begin{tblr}{
  width = \linewidth,
  colspec = {Q[90]Q[85]Q[72]Q[72]Q[72]Q[75]Q[75]Q[85]Q[75]Q[85]Q[60]Q[60]Q[60]},
  cell{1}{1} = {r=2}{},
  cell{1}{2} = {c=3}{},
  cell{1}{5} = {c=3}{},
  cell{1}{8} = {c=3}{},
  cell{1}{11} = {c=3}{},
  vlines,
  hline{1,3-7} = {-}{},
  hline{2} = {2-13}{},
}
Method           & CIFAR-100                                           &                                                    &                                                    & TinyImageNet                                       &                                                     &                                                    & ImageNet-Subset                                     &                 &                                                      & ImageNet-1K &      &      \\
                 & T=5                                                 & T=10                                               & T=20                                               & T=5                                                & T=10                                                & T=20                                               & T=5                                                 & T=10            & T=20                                                 & T=5         & T=10 & T=20 \\
PASS  *          & 25.20                                               & 30.25                                              & 30.61                                              & 18.04                                              & 23.11                                               & 30.55                                              & 19.66                                               & 25.85           & 30.98                                                & -           & -    & -    \\
IL2A             & 28.72                                               & 39.86                                              & 40.70                                              & 19.74                                              & 29.90                                               & 39.99                                              & -                                                   & -               & -                                                    & -           & -    & -    \\
SSRE *           & 18.37                                               & 19.48                                              & 19.00                                              & 9.17                                               & 14.06                                               & 14.20                                              & -                                                   & \textbf{8.30}   & -                                                    & -           & -    & -    \\
\textbf{NAPA-VQ} &{\textbf{6.90}\\\textbf{\textcolor{red}{\small{(-11.47)}}}}& {\textbf{9.65}\\\textbf{\textcolor{red}{\small{(-9.83)}}}}& {\textbf{9.08}\\\textbf{\textcolor{red}{\small{(-9.92)}}}}& {\textbf{9.08}\\\textbf{\textcolor{red}{\small{(-0.09)}}}} & {\textbf{10.81}\\\textbf{\textcolor{red}{\small{(-3.25)}}}} & {\textbf{9.31}\\\textbf{\textcolor{red}{\small{(-4.89)}}}} & {\textbf{7.17}\\\textbf{\textcolor{red}{\small{(-12.49)}}}} & {9.67\\\small{(+1.37)}} & {\textbf{14.49}\\\textbf{\textcolor{red}{\small{(-16.49)}}}} & \textbf{10.45}        & \textbf{10.94} & \textbf{18.23} 
\end{tblr}
\end{table*}

\begin{table*}[t]
\centering
\caption{Average Accuracy (\%) and Average Forgetting (\%) obtained for the ablation study conducted using CIFAR-100 to evaluate the effectiveness of NA-VQ and NA-PA. We refer to the combination of DCE and NA as NA-VQ.}
\label{Tab:ablation}
\begin{tblr}{
  width = \linewidth,
  colspec = {Q[392]Q[88]Q[88]Q[88]Q[90]Q[90]Q[90]},
  cell{1}{1} = {r=2}{},
  cell{1}{2} = {c=3}{0.264\linewidth},
  cell{1}{5} = {c=3}{0.27\linewidth},
  vlines,
  hline{1,3-8} = {-}{},
  hline{2} = {2-7}{},
}
Ablation~                                       & Average Accuracy $\uparrow$ &                &                & Average Forgetting $\downarrow$ &               &                \\
                                                & T=5              & T=10           & T=20           & T=5                & T=10          & T=20           \\
1) KD + CCE (Baseline)                          & 26.21            & 16.70          & 10.93          & 85.29              & 89.77         & 93.49          \\
2) KD + \textbf{DCE} & 49.83            & 32.90          & 17.01          & 48.84              & 75.37         & 81.71          \\
3) KD + DCE + \textbf{NA} $\rightarrow$ KD + \textbf{NA-VQ}           & 49.31            & 42.16          & 42.13          & 45.03              & 41.76         & 30.09          \\
4) KD + NA-VQ + \textbf{Gaussian-PA}                & 68.84            & 65.44          & 62.39          & 10.78              & 16.20         & 18.30          \\
5) KD + NA-VQ + \textbf{NA-PA} $\rightarrow$ \textbf{NAPA-VQ}              & \textbf{70.44}   & \textbf{69.04} & \textbf{67.42} & \textbf{ 6.90}     & \textbf{ 9.65} & \textbf{ 9.08} 
\end{tblr}

\end{table*}

\begin{figure}[t]
\centering
\includegraphics[width=0.42\textwidth]{
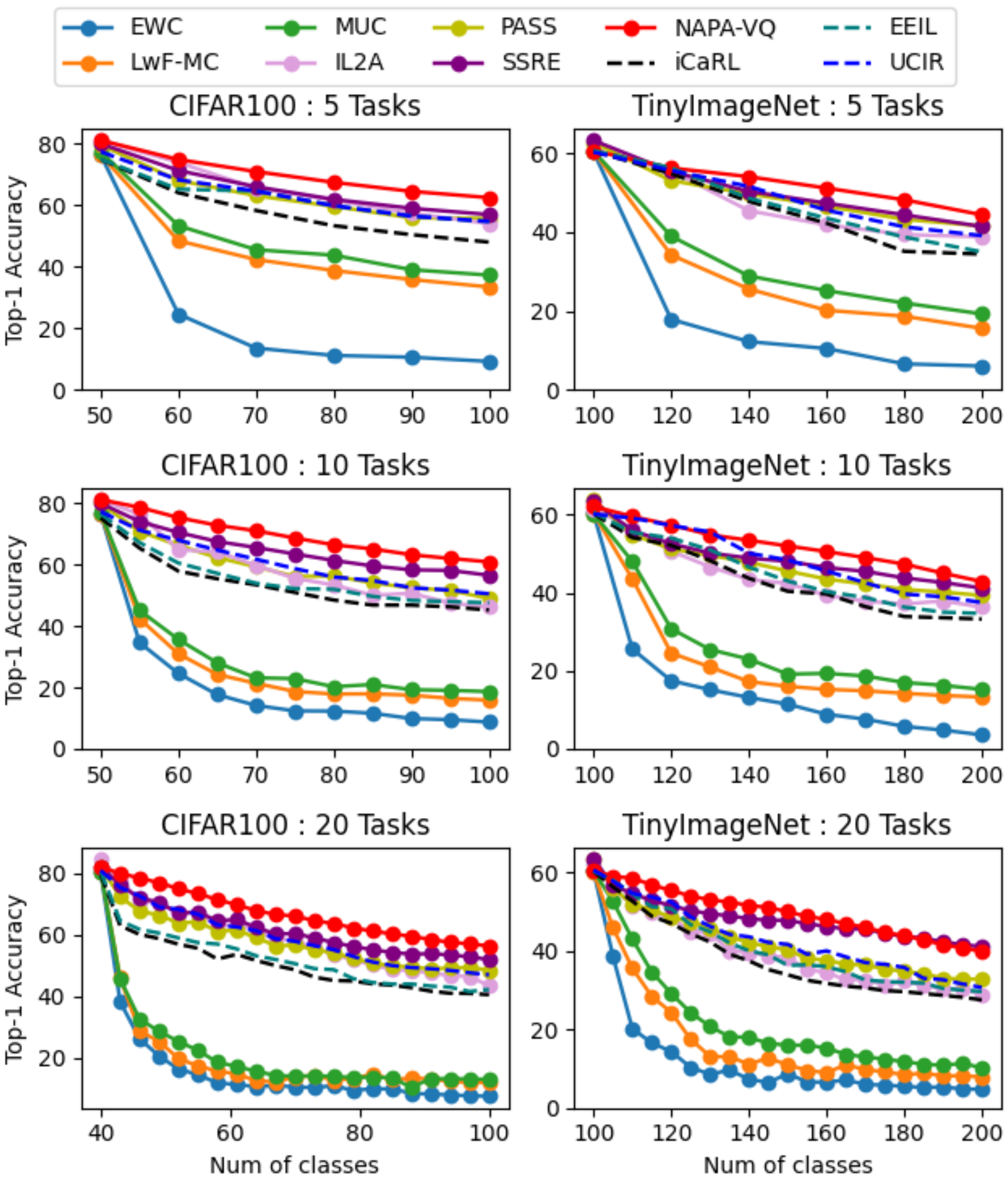}
\caption{Detailed Accuracy curves showing the Top-1 Accuracy at each incremental step.}
\label{fig:acc}
\end{figure} 

\subsection{Evaluation metrics}
In line with previous works \cite{Zhu2021PrototypeLearning,Zhu2022Self-SustainingLearning}, we report the standard metrics used to evaluate CIL strategies: Average Accuracy and Average Forgetting. Accuracy at task $t$ is the average accuracy of all the classes that have been learned up to and during task $t$. 
\textbf{Average accuracy} \cite{Chaudhry2018RiemannianIntransigence} is the mean accuracy across all the tasks, including the initial task. \textbf{Forgetting} at any given time for a task previously encountered, is measured by the difference between the maximum accuracy for the task during the learning process and the current accuracy for the same task. \textbf{Average Forgetting} at the end of task $t$ is therefore defined as the average of forgetting values for all the tasks learned up to task $t$ \cite{Chaudhry2018RiemannianIntransigence}. 
We report the average forgetting at the end of the final task. Additional explanations related to evaluation metrics can be found in Supp. materials.

\begin{figure}[t]
\centering
\includegraphics[width=0.47\textwidth]{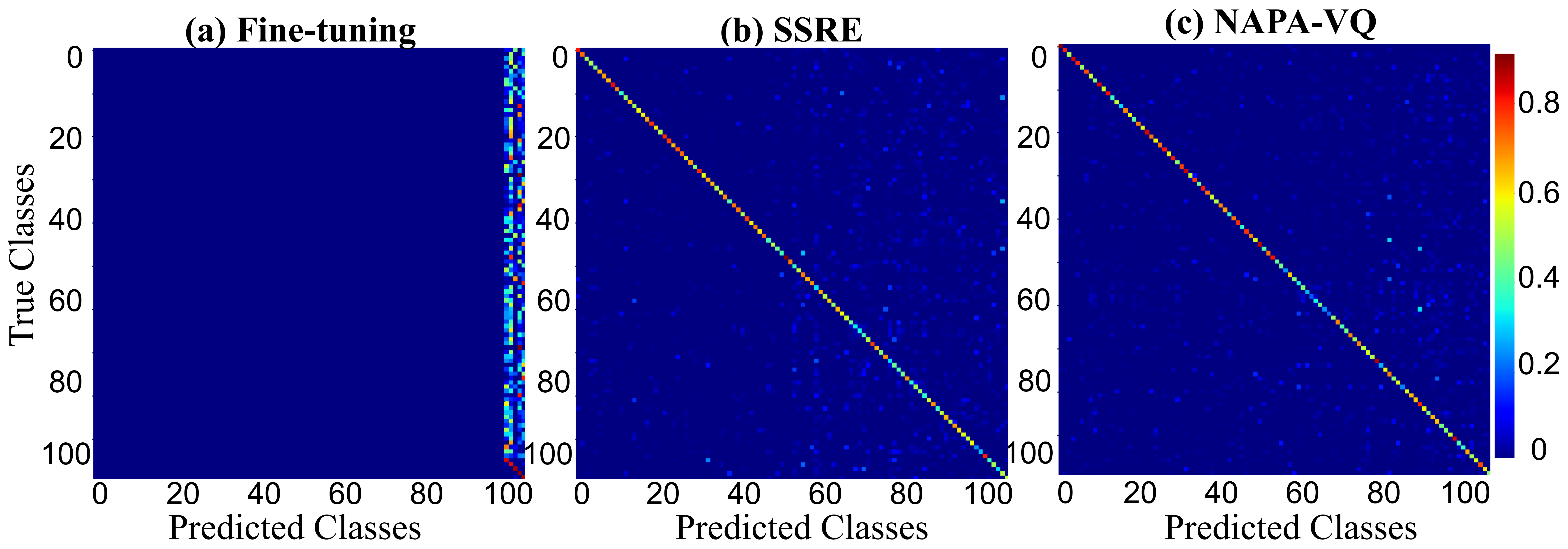}
\caption{Confusion matrices for Fine-tuning, SSRE and NAPA-VQ for CIFAR-100. Both SSRE and NAPA-VQ reduce the task recency bias observed in fine-tuning. Along the diagonal, NAPA-VQ has more red patches than SSRE.}
\label{fig:cfm}
\end{figure}

\begin{figure*}[t]
\centering
\includegraphics[width=0.98\textwidth]{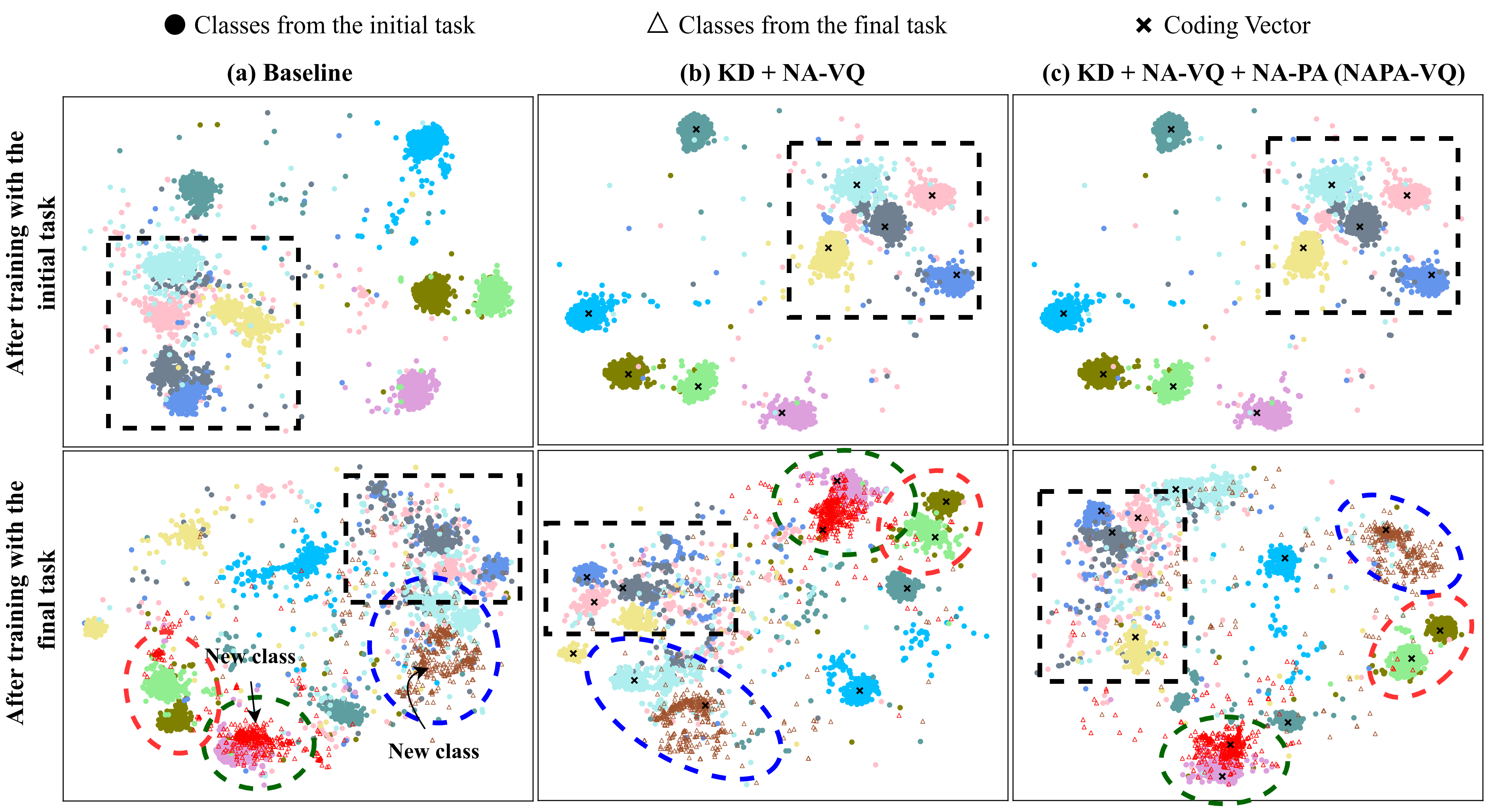}
\caption{Visualization of the impact of NA-VQ and NA-PA on the feature representations. Each colour represents a single class. The areas highlighted depict the observable differences between experiments. After learning the initial task (first row), NA-VQ integrated models (b and c) reduces much of the class overlap seen in the baseline model (a). The feature space for (b) and (c) are identical at this stage since NA-PA is only applied to the incremental tasks. After learning the final task (second row), NA-VQ integrated model (b) shows better discrimination between the old classes as well as between old and new classes compared to the baseline (a). When NA-PA is integrated, the discrimination between old and new classes improves further. 
}
\label{fig:overlap}
\end{figure*} 
\subsection{Comparison with SOTA}

We compare NAPA-VQ with the existing state-of-the-art (SOTA) methods in NECIL, including EWC \cite{Kirkpatrick2017OvercomingNetworks}, LwF\_MC \cite{Lampert2017}, MUC \cite{Liu2020MoreLearning},
% SDC \cite{Yu2020}, 
PASS \cite{Zhu2021PrototypeLearning}, IL2A \cite{Zhu2021Class-IncrementalAugmentation}, and SSRE \cite{Zhu2022Self-SustainingLearning}. We report the average accuracy and average forgetting of NAPA-VQ against the top three performing methods in Tables \ref{Tab:accuracy} and \ref{Tab:forgetting}. Reported values are the average of three separate runs. As illustrated in Table \ref{Tab:accuracy}, NAPA-VQ demonstrates an average improvement of 5\%, 2\%, and 4\% in accuracy for CIFAR-100, TinyImageNet, and ImageNet-Subset, respectively, over the best existing NECIL-SOTA technique. The detailed accuracy curves for the compared methods for CIFAR-100, TinyImageNet (in Fig. \ref{fig:acc}) and ImageNet-Subset (in Supp. Fig. 1), show that NAPA-VQ maintains higher accuracies over the incremental tasks. Moreover, NAPA-VQ exhibits a significant reduction in forgetting by an average of 10\%, 3\% and 9\% for CIFAR-100, TinyImageNet, and ImageNet-Subset, respectively. This reduction in forgetting is prominent when dealing with a larger number of tasks (Table \ref{Tab:forgetting}). 
In addition, we provide results on ImageNet-1K, demonstrating the effectiveness of NAPA-VQ on large-scale datasets. Furthermore, we compare NAPA-VQ to traditional exemplar-based methods, iCARL \cite{Lampert2017}, EEIL \cite{Castro2018End-to-endLearning}, and UCIR \cite{Hou2019LearningRebalancing} trained using a limited number of exemplars (20) (Fig. \ref{fig:acc}) and show that NAPA-VQ obtains competitive performance.  

\subsection{Ablation study}
     The ablation study is conducted using the CIFAR-100 dataset to demonstrate the impact of NA-VQ and NA-PA (Table \ref{Tab:ablation}). 
     % The success of our method can be attributed to our vector quantization mechanism ($SuNG\_VQ$) and the prototype generation mechanism ($SuNG\_PA$). 
     We first train a baseline model using the Categorical Cross Entropy loss ($L_{CCE}$) and $L_{KD}$ (Ablation-1) \cite{Zhu2022Self-SustainingLearning}. We then incorporate NA-VQ (Ablation-2 and 3) and NA-PA (Ablation-5) into the baseline model in sequence. For comparability, all the models use the same ResNet-18 architecture and the rotation-based data transformation.

  NA-VQ combines two losses; $L_{DCE}$ and $L_{NA}$ thus these losses are separately incorporated into the baseline model to comprehend their individual impact. First we substitute $L_{DCE}$ for $L_{CCE}$ (Ablation-2). Although this replacement improves accuracy across all three incremental scenarios compared to the baseline, the forgetting is still significant when $T=10$ and $20$. In Ablation-3, we add $L_{NA}$ to $L_{DCE}$ and $L_{KD}$ and observe comparable or better accuracy in all three scenarios while significantly reducing forgetting compared to  Ablation-2. This showcases the effectiveness of $L_{NA}$ in mitigating the interference of feature representations over the incremental steps. Finally, we integrate NA-PA in Ablation-5 and compare results to Ablation-3 which does not employ any prototypes and Ablation-4 which employs Gaussian-augmented prototypes  \cite{Zhu2021PrototypeLearning}. Our findings show that NA-PA enhances accuracy and reduces forgetting in all incremental scenarios. This improvement can be attributed to the prototypes generated closer to the boundaries which aid in identifying optimal decision boundaries, subsequently reducing the misclassification rate. Moreover, the advantages of NA-VQ and NA-PA are prominent when a larger number of tasks are involved, highlighting the method's importance.
\subsection{Reduced overlapping in the feature space}

We visualize the feature space of $T=10$ scenario in Ablation-1 (Baseline), Ablation-3 (KD + NA-VQ) and Ablation-5  (NAPA-VQ) in Fig. \ref{fig:overlap} using t-SNE \cite{derMaaten2008VisualizingT-SNE.} to show the impact of NA-VQ and NA-PA. Specifically, we visualize the feature representations of a randomly selected subset of classes learned during the initial task at two-time points: (1) after the initial task training, and (2) after the final task training with a subset of new classes from the final task. Once the first task is learned, the class-representative features of the baseline model overlap, whereas those of the NA-VQ integrated models are more compact and distinct, reducing the misclassification rate. Once the final task is learned, the overlapping in the baseline model increases creating further confusion between classes. NA-VQ integrated model reduces this overlap due to the more discretized feature space of the old classes and the repulsive forces between old and new classes. When NA-PA is integrated on top of NA-VQ the discrimination between the old classes as well as the discrimination between the old and new classes improve further showing the positive impact of prototypes.

\subsection{Comparison of confusion matrices}

Fig. \ref{fig:cfm} shows a comparison between the confusion matrices generated for (1) simple Fine-tuning where a model is trained using CCE loss incrementally without using any strategies to mitigate forgetting, (2) SSRE and (3) NAPA-VQ. The diagonal entries in the matrices represent correct predictions, while off-diagonal entries denote misclassifications. The predictions of Fine-tuning are heavily biased towards the most recent tasks due to the forgetting of old classes. SSRE and NAPA-VQ eliminate much of this bias by correctly classifying both old and new classes. Although quite similar, more red patches are visible along the diagonal in NAPA-VQ compared to SSRE, which explains the higher average accuracies in NAPA-VQ compared to SSRE.

%-------------------------------------------------------------------------
\subsection{Impact of the connectivity factor $K$}

 To determine the impact of the connectivity factor $K$ on performance, we conducted an experiment on the CIFAR-100 dataset by varying the value of $K$ between 2 and 50, with $K = 2$ being the commonly used heuristic \cite{ThomasMartinetzandKlausSchulten1991ATopologies, Fritzke1994ATopologies}. The results show that as $K$ increases, the performance improves but so does the running time (Supp. Fig. 3). The improved performance 
 can be attributed to a wider neighborhood being considered to
improve both decision boundary learning and prototype augmentation. A value of $ K = 15$ was found to provide desirable performance without compromising algorithm efficiency.

\section{Conclusion}
In this manuscript, we proposed NAPA-VQ, a novel method for CIL that does not rely on previous task exemplars to retain old knowledge. Instead, we increase the discriminability of the feature space by using class neighborhood information captured by a topological approximation of the feature space. Furthermore, we show that generating representative prototypes for old classes by borrowing the shared features of their neighboring classes helps to establish good decision boundaries between the areas where the classes tend to overlap. Comprehensive experiments on four benchmarking datasets demonstrate the superiority of our method over existing NECIL methods. While the proposed method exerts no limit on how many CVs to be used per class, we used one CV per class in our experiments. A future study may explore the effect of using a larger number of CVs per class on both the running time and the incremental learning performance. 

\textbf{Acknowledgements} T.M. acknowledges Melbourne Graduate Research Scholarship and
GCI Women in STEM Student Award support scheme. D.S. and S.H.
acknowledge Australian Research Council grant DP210101135. The authors thank Sachith Seneviratne, Maneesha Perera, Rashindrie Perera, and Nisal Ranasinghe for proofreading.

{\small
\bibliographystyle{ieee_fullname}
\bibliography{references}

\begin{thebibliography}{10}\itemsep=-1pt

\bibitem{Aljundi2018MemoryForget}
Rahaf Aljundi, Francesca Babiloni, Mohamed Elhoseiny, Marcus Rohrbach, and
  Tinne Tuytelaars.
\newblock {Memory Aware Synapses: Learning What (not) to Forget}.
\newblock In {\em Proceedings of the European conference on computer vision
  (ECCV)}, pages 139--154, 2018.

\bibitem{Aljundi2017ExpertExperts}
Rahaf Aljundi, Punarjay Chakravarty, and Tinne Tuytelaars.
\newblock {Expert gate: Lifelong learning with a network of experts}.
\newblock In {\em Proceedings of the IEEE Conference on Computer Vision and
  Pattern Recognition}, pages 3366--3375, 2017.

\bibitem{Aljundi2019GradientLearning}
Rahaf Aljundi, Min Lin, Baptiste Goujaud, and Yoshua Bengio.
\newblock {Gradient based sample selection for online continual learning}.
\newblock {\em Advances in neural information processing systems}, 32, 2019.

\bibitem{Blaes2017Few-shotPrototyping}
Sebastian Blaes and Thomas Burwick.
\newblock {Few-shot learning in deep networks through global prototyping}.
\newblock {\em Neural Networks}, 94:159--172, 10 2017.

\bibitem{Castro2018End-to-endLearning}
Francisco~M. Castro, Manuel~J. Mar{\'{i}}n-Jim{\'{e}}nez, Nicolás Guil,
  Cordelia Schmid, and Karteek Alahari.
\newblock {End-to-end incremental learning}.
\newblock In {\em Proceedings of the European conference on computer vision
  (ECCV)}, volume~CS, pages 233--248, 2018.

\bibitem{Chaudhry2018RiemannianIntransigence}
Arslan Chaudhry, Puneet~K. Dokania, Thalaiyasingam Ajanthan, and Philip~H.S.
  Torr.
\newblock {Riemannian Walk for Incremental Learning: Understanding Forgetting
  and Intransigence}.
\newblock In {\em Proceedings of the European conference on computer vision
  (ECCV)}, pages 556--572, 2018.

\bibitem{Chaudhry2019EfficientA-gem}
Arslan Chaudhry, Marc~Aurelio Ranzato, Marcus Rohrbach, and Mohamed Elhoseiny.
\newblock {Efficient lifelong learning with a-gem}.
\newblock In {\em 7th International Conference on Learning Representations,
  ICLR 2019}, pages 1--20, 2019.

\bibitem{Chaudhry2019OnLearning}
Arslan Chaudhry, Marcus Rohrbach, Mohamed Elhoseiny, Thalaiyasingam Ajanthan,
  Puneet~K Dokania, Philip H~S Torr, and Marc'Aurelio Ranzato.
\newblock {On tiny episodic memories in continual learning}.
\newblock {\em arXiv preprint arXiv:1902.10486}, 2019.

\bibitem{Chen2021IncrementalSpace}
Kuilin Chen and Chi-guhn Lee.
\newblock {Incremental Few-shot Learning via Vector Quantization in deep
  embedded space}.
\newblock In {\em International Conference on Learning Representations}, 2021.

\bibitem{Chu2020FeatureData}
Peng Chu, Xiao Bian, Shaopeng Liu, and Haibin Ling.
\newblock {Feature Space Augmentation for Long-Tailed Data}.
\newblock In {\em Computer Vision–ECCV}, pages 694--710, 2020.

\bibitem{deVen2018GenerativeLearning}
Gido~M de Ven and Andreas~S Tolias.
\newblock {Generative replay with feedback connections as a general strategy
  for continual learning}.
\newblock {\em arXiv preprint arXiv:1809.10635}, 2018.

\bibitem{DeVries2016DeepQuantization}
Harm De~Vries, Roland Memisevic, and Aaron Courville.
\newblock {Deep learning vector quantization}.
\newblock {\em ESANN 2016 - 24th European Symposium on Artificial Neural
  Networks}, (April):503--508, 2016.

\bibitem{DeLange2021ATasks}
Matthias Delange, Rahaf Aljundi, Marc Masana, Sarah Parisot, Xu Jia, Ales
  Leonardis, Greg Slabaugh, and Tinne Tuytelaars.
\newblock {A continual learning survey: Defying forgetting in classification
  tasks}.
\newblock {\em IEEE Transactions on Pattern Analysis and Machine Intelligence},
  8828(c):1--20, 2021.

\bibitem{Deng2009Imagenet:Database}
Jia Deng, Wei Dong, Richard Socher, Li-Jia Li, Kai Li, and Li Fei-Fei.
\newblock {Imagenet: A large-scale hierarchical image database}.
\newblock In {\em 2009 IEEE conference on computer vision and pattern
  recognition}, pages 248--255, 2009.

\bibitem{derMaaten2008VisualizingT-SNE.}
Laurens der Maaten and Geoffrey Hinton.
\newblock {Visualizing data using t-SNE.}
\newblock {\em Journal of machine learning research}, 9(11), 2008.

\bibitem{Dhar2019LearningMemorizing}
Prithviraj Dhar, Rajat~Vikram Singh, Kuan~Chuan Peng, Ziyan Wu, and Rama
  Chellappa.
\newblock {Learning without memorizing}.
\newblock {\em Proceedings of the IEEE Computer Society Conference on Computer
  Vision and Pattern Recognition}, 2019-June:5133--5141, 2019.

\bibitem{Fernando2017Pathnet:Networks}
Chrisantha Fernando, Dylan Banarse, Charles Blundell, Yori Zwols, David Ha,
  Andrei~A Rusu, Alexander Pritzel, and Daan Wierstra.
\newblock {Pathnet: Evolution channels gradient descent in super neural
  networks}.
\newblock {\em arXiv preprint arXiv:1701.08734}, 2017.

\bibitem{Fix1989DiscriminatoryProperties}
Evelyn Fix and J.~L. Hodges.
\newblock {Discriminatory Analysis. Nonparametric Discrimination: Consistency
  Properties}.
\newblock {\em International Statistical Review / Revue Internationale de
  Statistique}, 57(3):238, 12 1989.

\bibitem{Fritzke1994ATopologies}
Bernd Fritzke.
\newblock {A growing neural gas network learns topologies}.
\newblock {\em Advances in Neural Information Processing Systems}, 1994.

\bibitem{Girshick2015FastR-CNN}
Ross Girshick.
\newblock {Fast R-CNN}.
\newblock In {\em Proceedings of the IEEE international conference on computer
  vision.}, pages 1440--1448, 2015.

\bibitem{Hadsell2020EmbracingNetworks}
Raia Hadsell, Dushyant Rao, Andrei~A. Rusu, and Razvan Pascanu.
\newblock {Embracing Change: Continual Learning in Deep Neural Networks}.
\newblock {\em Trends in Cognitive Sciences}, 24(12):1028--1040, 12 2020.

\bibitem{Hammer2005SupervisedMeasure}
Barbara Hammer, Marc Strickert, and Thomas Villmann.
\newblock {Supervised neural gas with general similarity measure}.
\newblock {\em Neural Processing Letters}, 21(1):21--44, 2005.

\bibitem{He2016DeepRecognition}
Kaiming He, Xiangyu Zhang, Shaoqing Ren, and Jian Sun.
\newblock {Deep residual learning for image recognition}.
\newblock {\em Proceedings of the IEEE Computer Society Conference on Computer
  Vision and Pattern Recognition}, 2016-December:770--778, 12 2016.

\bibitem{Hou2019LearningRebalancing}
Saihui Hou, Xinyu Pan, Chen~Change Loy, Zilei Wang, and Dahua Lin.
\newblock {Learning a unified classifier incrementally via rebalancing}.
\newblock {\em Proceedings of the IEEE Computer Society Conference on Computer
  Vision and Pattern Recognition}, 2019-June:831--839, 2019.

\bibitem{Kirkpatrick2017OvercomingNetworks}
James Kirkpatrick, Razvan Pascanu, Neil Rabinowitz, Joel Veness, Guillaume
  Desjardins, Andrei~A. Rusu, Kieran Milan, John Quan, Tiago Ramalho, Agnieszka
  Grabska-Barwinska, Demis Hassabis, Claudia Clopath, Dharshan Kumaran, and
  Raia Hadsell.
\newblock {Overcoming catastrophic forgetting in neural networks}.
\newblock {\em Proceedings of the National Academy of Sciences of the United
  States of America}, 114(13):3521--3526, 2017.

\bibitem{Kohonen1990ImprovedQuantization}
Teuvo Kohonen.
\newblock {Improved versions of learning vector quantization}.
\newblock In {\em ijcnn international joint conference on Neural networks},
  pages 545--550. Publ by IEEE, 1990.

\bibitem{Kohonen1990TheMap}
Teuvo Kohonen.
\newblock {The Self-Organizing Map}.
\newblock {\em Proceedings of the IEEE}, 78(9):1464--1480, 1990.

\bibitem{Krizhevsky2009LearningImages}
Alex Krizhevsky, Geoffrey Hinton, and {others}.
\newblock {Learning multiple layers of features from tiny images}.
\newblock 2009.

\bibitem{Krizhevsky2012ImageNetNetworks}
Alex Krizhevsky, Ilya Sutskever, and Geoffrey~E. Hinton.
\newblock {ImageNet Classification with Deep Convolutional Neural Networks}.
\newblock {\em Advances in Neural Information Processing Systems}, 25, 2012.

\bibitem{Lafayette2016SpartanChimera}
Lev Lafayette, Greg Sauter, Linh Vu, and Bernard Meade.
\newblock {Spartan performance and flexibility: An hpc-cloud chimera}.
\newblock {\em OpenStack Summit, Barcelona}, 27, 2016.

\bibitem{Le2015TinyChallenge}
Ya Le and Xuan Yang.
\newblock {Tiny imagenet visual recognition challenge}.
\newblock {\em CS 231N}, 7(7):3, 2015.

\bibitem{Li2018LearningForgetting}
Zhizhong Li and Derek Hoiem.
\newblock {Learning without Forgetting}.
\newblock {\em IEEE Transactions on Pattern Analysis and Machine Intelligence},
  40(12):2935--2947, 2018.

\bibitem{Liu2018RotateForgetting}
Xialei Liu, Marc Masana, Luis Herranz, Joost Van De~Weijer, Antonio~M. Lopez,
  and Andrew~D. Bagdanov.
\newblock {Rotate your Networks: Better Weight Consolidation and Less
  Catastrophic Forgetting}.
\newblock {\em Proceedings - International Conference on Pattern Recognition},
  2018-August:2262--2268, 11 2018.

\bibitem{Liu2020MoreLearning}
Yu Liu, Sarah Parisot, Gregory Slabaugh, Xu Jia, Ales Leonardis, and Tinne
  Tuytelaars.
\newblock {More Classifiers, Less Forgetting: A Generic Multi-classifier
  Paradigm for Incremental Learning}.
\newblock In {\em Computer Vision–ECCV 2020: 16th European Conference}, pages
  699--716, 2020.

\bibitem{Long2015FullySegmentation}
Jonathan Long, Evan Shelhamer, and Trevor Darrell.
\newblock {Fully Convolutional Networks for Semantic Segmentation}.
\newblock In {\em Proceedings of the IEEE conference on computer vision and
  pattern recognition}, pages 3431--3440, 2015.

\bibitem{NIPS2017_f8752278}
David Lopez-Paz and Marc\textquotesingle~Aurelio Ranzato.
\newblock {Gradient Episodic Memory for Continual Learning}.
\newblock In I Guyon, U~Von Luxburg, S Bengio, H Wallach, R Fergus, S
  Vishwanathan, and R Garnett, editors, {\em Advances in Neural Information
  Processing Systems}, volume~30. Curran Associates, Inc., 2017.

\bibitem{Mallya2018Packnet:Pruning}
Arun Mallya and Svetlana Lazebnik.
\newblock {Packnet: Adding multiple tasks to a single network by iterative
  pruning}.
\newblock In {\em Proceedings of the IEEE conference on Computer Vision and
  Pattern Recognition}, pages 7765--7773, 2018.

\bibitem{Lampert2017}
Sylvestre~Alvise Rebuffi, Alexander Kolesnikov, Georg Sperl, and Christoph~H.
  Lampert.
\newblock {iCaRL: Incremental classifier and representation learning}.
\newblock {\em Proceedings - 30th IEEE Conference on Computer Vision and
  Pattern Recognition, CVPR 2017}, 2017-Janua:5533--5542, 2017.

\bibitem{Rusu2016ProgressiveNetworks}
Andrei~A Rusu, Neil~C Rabinowitz, Guillaume Desjardins, Hubert Soyer, James
  Kirkpatrick, Koray Kavukcuoglu, Razvan Pascanu, and Raia Hadsell.
\newblock {Progressive neural networks}.
\newblock {\em arXiv preprint arXiv:1606.04671}, 2016.

\bibitem{Sato1996GeneralizedQuantization}
Atsushi Sato and Keiji Yamada.
\newblock {Generalized Learning Vector Quantization}.
\newblock {\em Advances in neural information processing systems}, pages
  423--429, 1996.

\bibitem{Seff2017ContinualNets}
Ari Seff, Alex Beatson, Daniel Suo, and Han Liu.
\newblock {Continual learning in generative adversarial nets}.
\newblock {\em arXiv preprint arXiv:1705.08395}, 2017.

\bibitem{Serra2018OvercomingTask}
Joan Serra, Didac Suris, Marius Miron, and Alexandros Karatzoglou.
\newblock {Overcoming catastrophic forgetting with hard attention to the task}.
\newblock In {\em International Conference on Machine Learning}, pages
  4548--4557, 2018.

\bibitem{Shin2017}
Hanul Shin, Jung~Kwon Lee, Jaehong Kim, and Jiwon Kim.
\newblock {Continual learning with deep generative replay}.
\newblock In {\em Advances in Neural Information Processing Systems},
  volume~30, 2017.

\bibitem{Smith2021AlwaysLearning}
James Smith, Yen-Chang Hsu, Jonathan Balloch, Yilin Shen, Hongxia Jin, and
  Zsolt Kira.
\newblock {Always be dreaming: A new approach for data-free class-incremental
  learning}.
\newblock In {\em Proceedings of the IEEE/CVF International Conference on
  Computer Vision}, pages 9374--9384, 2021.

\bibitem{Smith2022ALearning}
James~Seale Smith, Junjiao Tian, Yen-Chang Hsu, and Zsolt Kira.
\newblock {A Closer Look at Rehearsal-Free Continual Learning}.
\newblock (2019):1--12, 2022.

\bibitem{Tao2020Topology-PreservingClassificatio}
Xiaoyu Tao, Xinyuan Chang, Xiaopeng Hong, and Xing Wei.
\newblock {Topology-Preserving Class-Incremental Learning for Image
  Classificatio}.
\newblock {\em European Conference on Computer Vision}, (Cil), 2020.

\bibitem{Tao2020Few-ShotLearning}
Xiaoyu Tao, Xiaopeng Hong, Xinyuan Chang, Songlin Dong, Xing Wei, and Yihong
  Gong.
\newblock {Few-Shot Class-Incremental Learning}.
\newblock {\em Proceedings of the IEEE Computer Society Conference on Computer
  Vision and Pattern Recognition}, pages 12180--12189, 2020.

\bibitem{Thanh-Tung2020CatastrophicGans}
Hoang Thanh-Tung and Truyen Tran.
\newblock {Catastrophic forgetting and mode collapse in gans}.
\newblock In {\em 2020 international joint conference on neural networks
  (ijcnn)}, pages 1--10, 2020.

\bibitem{ThomasMartinetzandKlausSchulten1991ATopologies}
{Thomas Martinetz and Klaus Schulten}.
\newblock {A "Neural-Gas" Network Learns Topologies}.
\newblock {\em Artificial Neural Networks}, page 397–402, 1991.

\bibitem{Villmann2017FusionLearning}
T. Villmann, M. Biehl, A. Villmann, and S. Saralajew.
\newblock {Fusion of deep learning architectures, multilayer feedforward
  networks and learning vector quantizers for deep classification learning}.
\newblock {\em 12th International Workshop on Self-Organizing Maps and Learning
  Vector Quantization, Clustering and Data Visualization, WSOM 2017 -
  Proceedings}, 8 2017.

\bibitem{Yang2018RobustLearning}
Hong-ming Yang, Xu-yao Zhang, Fei Yin, and Cheng-lin Liu.
\newblock {Robust Classification with Convolutional Prototype Learning}.
\newblock In {\em Proceedings of the IEEE conference on Computer Vision and
  Pattern Recognition}, pages 3474--3482, 2018.

\bibitem{Yoon2017LifelongNetworks}
Jaehong Yoon, Eunho Yang, Jeongtae Lee, and Sung~Ju Hwang.
\newblock {Lifelong learning with dynamically expandable networks}.
\newblock {\em arXiv preprint arXiv:1708.01547}, 2017.

\bibitem{Yu2020}
Lu Yu, Bartłomiej Twardowski, Xialei Liu, Luis Herranz, Kai Wang, Yongmei
  Cheng, Shangling Jui, and Joost van~de Weijer.
\newblock {Semantic drift compensation for class-incremental learning}.
\newblock {\em Proceedings of the IEEE Computer Society Conference on Computer
  Vision and Pattern Recognition}, pages 6980--6989, 2020.

\bibitem{Zenke2017ContinualIntelligence}
Friedemann Zenke, Ben Poole, and Surya Ganguli.
\newblock {Continual learning through synaptic intelligence}.
\newblock In {\em 34th International Conference on Machine Learning, ICML
  2017}, volume~8, pages 6072--6082, 2017.

\bibitem{Zhang2020Class-incrementalConsolidation}
Junting Zhang, Jie Zhang, Shalini Ghosh, Dawei Li, Serafettin Tasci, Larry
  Heck, Heming Zhang, and C.-C.~Jay Kuo.
\newblock {Class-incremental Learning via Deep Model Consolidation}.
\newblock In {\em Proceedings of the IEEE/CVF Winter Conference on Applications
  of Computer Vision (WACV)}, 3 2020.

\bibitem{Zhu2021Class-IncrementalAugmentation}
Fei Zhu, Zhen Cheng, Xu-Yao Zhang, and Cheng-Lin Liu.
\newblock {Class-Incremental Learning via Dual Augmentation}.
\newblock In {\em Advances in Neural Information Processing Systems}, number
  NeurIPS, pages 14306--14318, 2021.

\bibitem{Zhu2021PrototypeLearning}
Fei Zhu, Xu-yao Zhang, Chuang Wang, Fei Yin, and Cheng-lin Liu.
\newblock {Prototype Augmentation and Self-Supervision for Incremental
  Learning}.
\newblock In {\em IEEE Computer Society Conference on Computer Vision and
  Pattern Recognition}, pages 5267--5876, 2021.

\bibitem{Zhu2022Self-SustainingLearning}
Kai Zhu, Wei Zhai, Yang Cao, Jiebo Luo, and Zheng-Jun Zha.
\newblock {Self-Sustaining Representation Expansion for Non-Exemplar
  Class-Incremental Learning}.
\newblock In {\em IEEE Computer Society Conference on Computer Vision and
  Pattern Recognition}, pages 9296--9305, 2022.

\end{thebibliography}
}

\end{document}